\documentclass[10pt,journal,onecolumn]{IEEEtran}

\usepackage{comment}
\usepackage{graphicx}
\usepackage{amsmath,amsfonts}
\usepackage{booktabs}
\usepackage{url}
\usepackage{hyperref}
\usepackage{xcolor}

\begin{document}

\title{A Usable GAN-Based Tool for Synthetic ECG Generation in Cardiac Amyloidosis Research}

\author{Speziale Francesco, Ugo Lomoio, Fabiola Boccuto, Pierangelo Veltri, and Pietro Hiram Guzzi%
\thanks{Affiliation, address, email.}}

\maketitle

\begin{abstract}
Cardiac amyloidosis (CA) is a rare and underdiagnosed infiltrative cardiomyopathy, and available datasets for machine-learning models are typically small, imbalanced and heterogeneous. This paper presents a Generative Adversarial Network (GAN) and a graphical command-line interface for generating realistic synthetic electrocardiogram (ECG) beats to support early diagnosis and patient stratification in CA. The tool is designed for usability, allowing clinical researchers to train class-specific generators once and then interactively produce large volumes of labelled synthetic beats that preserve the distribution of minority classes.
\end{abstract}


\section{Introduction}
Cardiac amyloidosis results from extracellular deposition of misfolded amyloid fibrils, most commonly immunoglobulin light chain (AL) or transthyretin (ATTR), within the myocardium\cite{castiglione2025ca_diagnosis_treatment} The disease leads to progressive restrictive cardiomyopathy, arrhythmias and heart failure, yet early clinical manifestations are subtle and non-specific\cite{cheng2012ecg_ca_findings,utility_ecg_amyloidosis} Artificial intelligence (AI) applied to standard 12‑lead ECG and echocardiography has recently demonstrated high diagnostic performance for CA, but current models are trained on limited cohorts collected at a few centres, often with strong imbalance between CA and non-CA patients \cite{mayo_ai_ecg_amyloidosis,ai_ecg_echocardiography_ca,schrutka2022ecg_ca}.

Data scarcity and imbalance are well-recognised obstacles in cardiac signal analysis, particularly for rare phenotypes and arrhythmias \cite{cecggan_imbalance_2024,autoencoder_synth_ecg_2024}. Traditional oversampling and undersampling often distort the data distribution or discard informative signals, whereas simple transformations of ECG (e.g., scaling, shifting) fail to reproduce realistic morphology \cite{cecggan_imbalance_2024,synthetic_ecg_2025}. Generative adversarial networks (GANs) can learn the underlying data distribution and generate synthetic samples that resemble real ECG beats, providing a powerful strategy to balance datasets and support downstream classification and clustering tasks \cite{arora2022gans_synthetic_patient,synthetic_ecg_2025,data_aug_gan_mitbih}.

This work describes a GAN-based tool for 1D ECG beat generation, originally developed on the MIT-BIH Arrhythmia Database\cite{mitbih_database} and designed for extension to CA-focused cohorts. The contribution is threefold: (1) a class-specific GAN architecture using bidirectional LSTMs for realistic beat synthesis; (2) a user-oriented workflow with separate training and generation scripts, emphasising reproducibility and ease of use; and (3) a discussion of how synthetic ECG beats can support clustering-based early diagnosis and patient grouping in CA.

\section{ECG Characteristics of Cardiac Amyloidosis}


Cardiac amyloidosis produces a distinctive but heterogeneous electrocardiographic phenotype that reflects diffuse myocardial infiltration, interstitial fibrosis and involvement of the conduction system \cite{cheng2012ecg_ca_findings,utility_ecg_amyloidosis}. Classical patterns include low QRS voltages, pseudoinfarction Q waves, atrioventricular (AV) and intraventricular conduction disturbances, repolarisation abnormalities and a high burden of atrial and ventricular arrhythmias \cite{electropatterns_arrhythmias_ca,ecg_meta_value_2024}.Although none of these findings is pathognomonic, their combination—particularly when discordant with marked left‑ventricular (LV) wall thickening on imaging—is strongly suggestive of CA and can trigger targeted diagnostic work‑up \cite{utility_ecg_amyloidosis,lv_mass_voltage_ratio_2020}.

\subsection{Low QRS voltage and voltage–mass discrepancy}
Low QRS voltage in the limb leads, typically defined as a peak‑to‑peak QRS amplitude \(\leq 0.5\)~mV in all limb leads, is one of the most frequently described ECG signs in CA \cite{cheng2012ecg_ca_findings,case_evolution_ecg_2021}. Cohort studies report low limb‑lead voltages in approximately 11–57\% of patients, with higher prevalence in AL than in ATTR forms, and low voltage is associated with greater myocardial amyloid burden and worse prognosis \cite{low_voltage_ca_2022,progression_ecg_ca_2024,ecg_meta_value_2024}. Many CA patients show increased LV wall thickness on echocardiography or cardiac magnetic resonance but fail to satisfy conventional ECG criteria for LV hypertrophy, leading to the concept of “paradoxical” LV hypertrophy and elevated LV mass–to–QRS voltage ratio \cite{lv_mass_voltage_ratio_2020,paradoxical_lvh_2023}. This voltage–mass discrepancy helps differentiate CA from hypertensive or hypertrophic cardiomyopathy and has been proposed as a screening and prognostic index in patients with otherwise unexplained LV hypertrophy \cite{paradoxical_lvh_2023,unexplained_lvh_prognosis_2021}.

\subsection{Pseudoinfarction patterns and repolarisation abnormalities}
Pseudoinfarction patterns, characterised by pathological Q waves or QS complexes in the absence of corresponding coronary artery disease, are another hallmark of CA \cite{cheng2012ecg_ca_findings,case_evolution_ecg_2021}. Cross‑sectional series indicate that 29–50\% of CA patients exhibit pseudoinfarction patterns, most commonly in the anterior and inferior precordial leads, reflecting subendocardial infiltration and regional conduction delay rather than transmural infarction \cite{pseudoinfarction_ca_2015,progression_ecg_ca_2024,ecg_meta_value_2024}. T‑wave inversion, ST‑segment depression and nonspecific repolarisation abnormalities are also frequent and may correlate with the extent of late gadolinium enhancement or extracellular volume expansion on cardiac magnetic resonance, suggesting that surface ECG can provide a low‑cost surrogate of myocardial amyloid burden when advanced imaging is unavailable\cite{ecg_vs_cmr_al_2021,lv_mass_voltage_ratio_2020}

\subsection{Conduction disturbances and rhythm disorders}
Amyloid infiltration of the conduction system predisposes to both AV nodal and intraventricular conduction disturbances \cite{utility_ecg_amyloidosis,electropatterns_arrhythmias_ca}. First‑degree AV block, bundle branch blocks and left anterior fascicular block are commonly observed, with large cohorts showing first‑degree AV block in roughly 23–40\% of patients and left bundle branch block more prevalent in ATTR wild‑type than in AL amyloidosis \cite{progression_ecg_ca_2024,electropatterns_arrhythmias_ca}. QRS duration tends to increase with disease progression and longer baseline QRS has been associated with more advanced diastolic dysfunction, higher amyloid stage and adverse outcomes, making it a potential marker of electrical disease severity \cite{progression_ecg_ca_2024,utility_ecg_amyloidosis}.

Supraventricular arrhythmias, particularly atrial fibrillation and atrial flutter, occur in 20–40\% of CA patients and contribute to thromboembolic risk and haemodynamic decompensation \cite{electropatterns_arrhythmias_ca,utility_ecg_amyloidosis}. Ventricular arrhythmias ranging from premature ventricular contractions to non‑sustained ventricular tachycardia are also frequent, especially in advanced disease, and may underlie a subset of sudden cardiac deaths \cite{electropatterns_arrhythmias_ca,ecg_meta_value_2024}. The high prevalence of rhythm disorders explains the frequent need for pacemaker implantation and complicates risk stratification for implantable cardioverter‑defibrillators in this population \cite{pacemaker_need_ca_2022,utility_ecg_amyloidosis}.

\subsection{Subtype‑specific and dynamic changes (AL vs ATTR)}
Electrocardiographic characteristics differ between AL and ATTR amyloidosis and evolve over time. Patients with AL amyloidosis more often display diffuse low QRS voltages, anterior and inferior pseudoinfarction patterns and widespread T‑wave inversion, reflecting rapid myocardial infiltration and extensive interstitial involvement \cite{low_voltage_ca_2022,electropatterns_arrhythmias_ca,progression_ecg_ca_2024}. In contrast, ATTR—especially wild‑type ATTR—typically shows higher QRS voltages, longer PR intervals, a higher prevalence of first‑degree AV block and left bundle branch block, and comparatively fewer low‑voltage patterns despite substantial LV wall thickening, pointing to a more indolent but conduction‑system‑centred disease course \cite{progression_ecg_ca_2024,utility_ecg_amyloidosis,ecg_meta_value_2024}. Longitudinal studies demonstrate that in both subtypes PR and QRS intervals prolong and voltage–mass discrepancy increases with disease progression, but these changes tend to occur earlier and more abruptly in AL, whereas ATTR cases show a slower, stepwise evolution toward advanced conduction disease \cite{progression_ecg_ca_2024,case_evolution_ecg_2021}. These subtype‑specific trajectories are clinically relevant for clustering and for GAN‑based synthetic augmentation, as generators conditioned on AL versus ATTR labels could capture and reproduce distinct ECG morphologies and progression patterns for downstream diagnostic and prognostic models \cite{ai_ecg_meta_analysis_2024,cecggan_imbalance_2024}.

\section{Clustering for Early Diagnosis and Patient Grouping}
Unsupervised learning and clustering have been applied to CA to uncover latent disease subtypes and to refine risk stratification beyond traditional staging schemes \cite{lomoio2025design,bargagna2023bayesian_ca_pet}. In ECG imaging, clustering of activation and voltage maps contributed to identifying patient groups with distinct patterns and supported derivation of simple criteria (e.g., voltage–mass indices) for screening \cite{schrutka2022ecg_ca}. Multimodal clustering combining ECG, echocardiography and biomarkers has also been proposed to separate CA from hypertrophic cardiomyopathy or hypertensive heart disease in populations with left ventricular hypertrophy \cite{ai_ecg_hcm_vs_ca_2024,ai_ecg_meta_analysis_2024}.

Clustering is particularly attractive for early CA detection because subtle ECG changes may precede overt imaging abnormalities and heart failure symptoms \cite{mayo_ai_ecg_amyloidosis,progression_ecg_ca_2024}. However, when the number of confirmed CA cases is small, standard clustering algorithms (e.g., \(k\)-means, Gaussian mixture models, spectral clustering) become unstable, and discovered clusters are sensitive to random initialisation and sampling noise \cite{bargagna2023bayesian_ca_pet,arora2022gans_synthetic_patient}. Synthetic ECG beats generated by GANs can help alleviate these limitations by augmenting CA-related patterns, improving density estimation in latent space and enabling more robust identification of clinically meaningful clusters for early and advanced disease, as well as AL versus ATTR phenotypes \cite{cecggan_imbalance_2024,synthetic_ecg_2025,data_aug_gan_mitbih}.

\section{Methods}
\subsection{Data source and preprocessing}

ECG beats were extracted from the MIT‑BIH Arrhythmia Database provided as training and test CSV files \cite{mitbih_database,project_train_gan} The two files were concatenated into a single dataframe, column 187 was renamed to \texttt{class}, and columns 0–186 contained 187 uniformly sampled points from a single beat.  Beat labels were mapped to five clinically meaningful classes: normal, atrial premature, premature ventricular contraction, fusion of ventricular and normal, and fusion of paced and normal; an additional column \texttt{label} stored the corresponding string label. 

The combined dataset was saved as \texttt{mitbih\_processed.csv} in a \texttt{data} directory, and class frequencies were computed and plotted as a colour‑coded bar chart, highlighting the strong predominance of normal beats (about 90\,589 samples, 82.77\%) and the scarcity of some fusion classes (e.g.\ 803 samples, 0.73\%)\cite{project_train_gan,data_dist_figure} For each class, a PyTorch \texttt{Dataset} accessed the relevant subset of rows, exposing ECG samples as single‑channel tensors of shape \((1,187)\) and the integer class as the target, while a \texttt{DataLoader} provided shuffled mini‑batches of size 96 for training.

\subsection{GAN architecture}
The augmentation pipeline is based on a standard GAN composed of a generator \(G\) and discriminator \(D\), both implemented in PyTorch and tailored to one‑dimensional ECG signals\cite{arora2022gans_synthetic_patient,synthetic_ecg_2025} The generator receives as input a noise tensor \(z \in \mathbb{R}^{B \times 1 \times 187}\) sampled from a unit normal distribution and produces synthetic ECG beats \(\hat{x} \in \mathbb{R}^{B \times 1 \times 187}\)  . Internally, \(z\) is processed by a bidirectional LSTM layer with input size 187, hidden size 128, one layer and batch‑first layout to model temporal dependencies along the beat; its output is reshaped to a vector of size 256 and passed through two fully connected layers (256 units each) with LeakyReLU activations and a dropout layer with \(p=0.2\) to mitigate overfitting  . A final linear layer with 187 outputs maps this latent representation to the time‑domain waveform, which is then reshaped back to \((1,187)\) to match the original beat format  .

The discriminator mirrors this design and acts as a binary classifier that estimates the probability that an input beat is real.  It comprises a bidirectional LSTM with input size 187 and hidden size 256, whose outputs are flattened to 512 features per beat and fed into two fully connected layers (512 and 256 units) with LeakyReLU activations and dropout, followed by a final sigmoid layer that outputs a scalar in \([0,1]\).  Both networks use Xavier‑style initialisation through PyTorch defaults, and all recurrent layers operate in batch‑first mode so that mini‑batches of shape \((B,1,187)\) can be efficiently processed on CPU or GPU. 

\subsection{Training strategy}
Training is performed separately for each arrhythmia class, yielding class‑specific generators that focus on reproducing the morphology and variability of a single beat type \cite{project_train_gan,project_generate_signals}. When the \texttt{train\_gan.py} script is executed, the user is shown the available classes and prompted to select the one used for training; the script then filters \texttt{mitbih\_processed.csv} accordingly and initialises the GAN components.  Before the training loop, a global random seed (2021) is set for NumPy and PyTorch to ensure experiment reproducibility, and the computation device is chosen automatically as CUDA (if available) or CPU. 

For each epoch, the trainer alternates between updating the discriminator and the generator using mini‑batches from the class‑specific \texttt{DataLoader}.  The discriminator loss \(\mathcal{L}_D\) is computed as binary cross‑entropy between its outputs and ground‑truth labels (1 for real beats, 0 for generated beats), while the generator loss \(\mathcal{L}_G\) is the binary cross‑entropy between discriminator predictions for generated beats and the target label 1, encouraging the generator to produce realistic signals \cite{project_train_gan,synthetic_ecg_2025}. Optimisation uses the Adam algorithm with a learning rate of \(2\times10^{-4}\) for both \(G\) and \(D\), and the training loop runs for 3\,000 epochs, logging discriminator and generator losses at each epoch and printing summaries every 300 epochs. 

At regular intervals the trainer generates a batch of synthetic beats from a fixed noise tensor and plots them to qualitatively monitor convergence and detect potential mode collapse or vanishing gradients \cite{project_train_gan,synthetic_ecg_2025} After training, the state dictionaries of \(G\) and \(D\) are saved in a \texttt{models} directory as \texttt{generator\_class\_<id>.pth} and \texttt{discriminator\_class\_<id>.pth}, respectively, allowing later re‑use without retraining \cite{project_train_gan,project_generate_signals}. In addition, the script saves publication‑ready figures of generator and discriminator loss curves and a side‑by‑side comparison of a real and synthetic beat for the trained class in a \texttt{results} directory. 

\subsection{Synthetic signal generation and export}
Once at least one class‑specific generator has been trained, synthetic beats can be produced using the dedicated \texttt{generate\_signals.py} script.. On start‑up, the script lists all classes and asks the user to choose the class to augment and the number of synthetic signals to generate; it then loads the corresponding generator weights from the \texttt{models} directory and switches the model to evaluation mode to disable dropout and gradient computation.. Synthetic beats are obtained by sampling independent noise tensors with shape \((N, 1, 187)\), passing them through the generator under a \texttt{no\_grad} context, and collecting the resulting array of shape \((N,187)\)..

The generated signals are wrapped in a pandas dataframe, with each row storing 187 samples plus two extra columns: the integer \texttt{class} and the human‑readable \texttt{label} corresponding to the selected beat type.. The dataframe is then written to a \texttt{synthetic\_data} directory as \texttt{synthetic\_<Label>.csv}; if a file with the same name already exists, the user can choose to append to the existing file or automatically create a new versioned file (\texttt{synthetic\_<Label>\_v2.csv}, \texttt{v3.csv}, and so on), which preserves previous experiments and supports traceability. The resulting CSV files mirror the structure of the original MIT‑BIH tables and can be directly merged with real beats for downstream training, evaluation or clustering analyses aimed at improving early detection and patient grouping in cardiac amyloidosis \cite{pcecggan_imbalance_2024}.

\section{Graphical User Interface}

\begin{figure}[t]
  \centering
  \includegraphics[width=\linewidth]{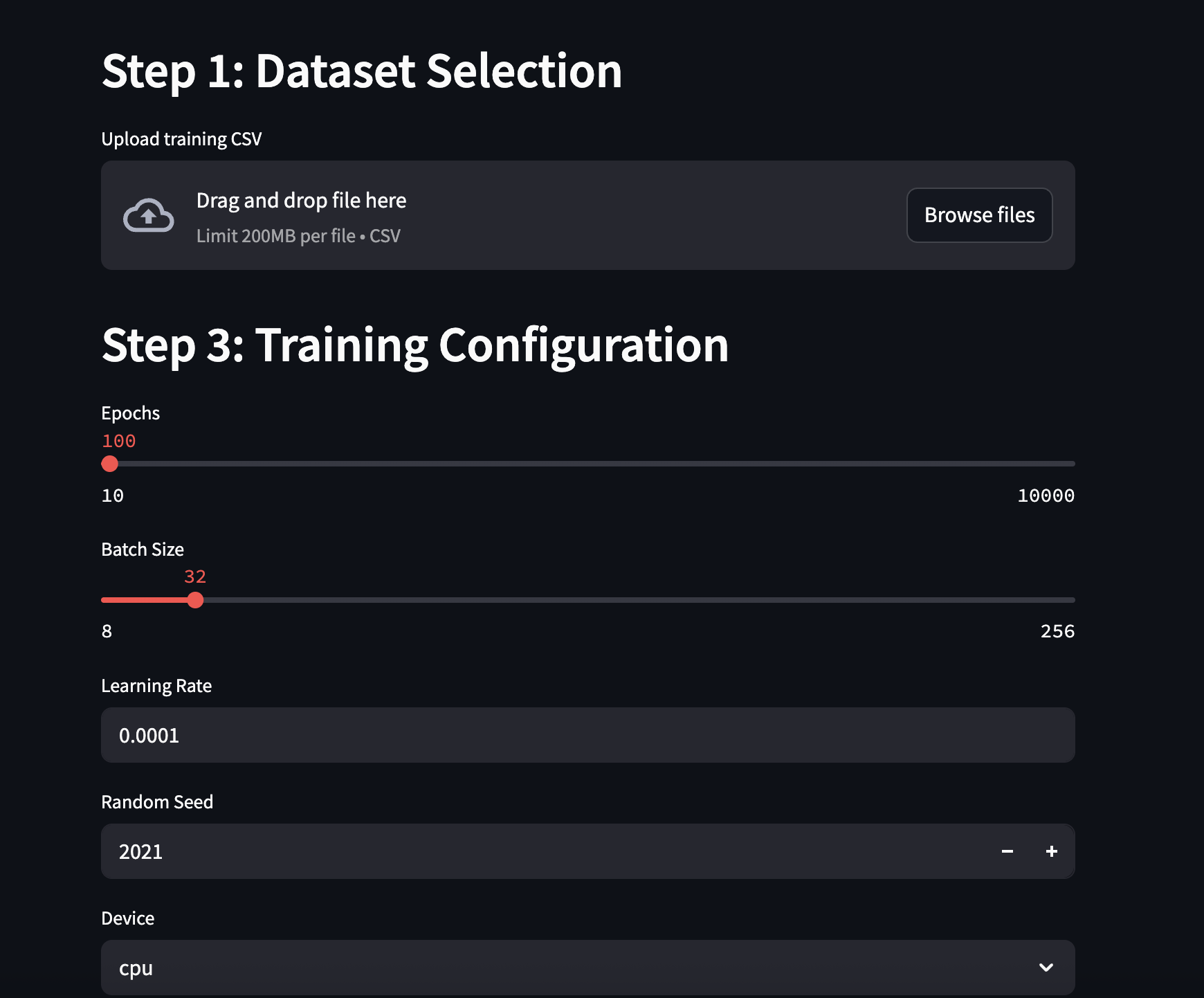}
  \caption{Streamlit-based graphical user interface for configuring GAN training. The panel illustrates dataset upload (Step~1) and hyper-parameter selection for epochs, batch size, learning rate, random seed and device (Step~3).}
  \label{fig:training_gui}
\end{figure}

The tool includes a \textbf{Streamlit}-based graphical user interface (GUI) that encapsulates the full training workflow for the GAN in an interactive web page, allowing users to configure experiments without editing Python code. The interface is organised into four numbered steps—dataset selection, target-class definition, training configuration and execution—mirroring the conceptual flow of the augmentation pipeline and providing immediate visual feedback on both input data and training progress (Fig.~\ref{fig:training_gui}). 

\subsection{Layout and interaction flow}
The main page displays the title \emph{Synthetic ECG GAN Trainer} and is divided into clearly separated sections implemented with \texttt{st.header} calls. 

\paragraph*{Step~1: Dataset selection}
A file-uploader widget (\texttt{st.file\_uploader}) accepts a CSV file containing ECG beats and loads it into a pandas dataframe; a tabular preview of the first rows is displayed by \texttt{st.write} so that users can verify column order and data format.  A drop-down menu allows the user to select the column containing class labels, and the unique values of this column are extracted to populate the list of available classes for subsequent steps.

\paragraph*{Step~2: Target class for augmentation}
If at least one class is detected, a second \texttt{selectbox} lets the user choose the target class on which the generator will be trained (e.g., ``Normal'', ``Premature ventricular contraction''). The selected value can be either numeric or string; in the latter case, it is translated into the internal class index through the \texttt{LABELS\_INV} dictionary imported from the training module, with explicit error messages if the label is not found. 

\paragraph*{Step~3: Training configuration}
Training hyper-parameters are exposed as intuitive controls, as illustrated in Fig.~\ref{fig:training_gui}.  The number of epochs and batch size are configured through sliders (10–10\,000 epochs, batch size 8–256), while the learning rate is entered as a text field with validation that enforces a positive numeric value and provides descriptive error messages for invalid input (e.g., non-numeric strings or non-positive values).  A numeric field sets the random seed, and a device selector lists the available options (CPU by default, with ``cuda'' automatically added when a GPU is detected via \texttt{torch.cuda.is\_available()}), facilitating reproducible experiments and straightforward exploitation of hardware acceleration.

\paragraph*{Step~4: Training execution and outputs}
When the \emph{Start Training} button is pressed, provided that a dataset and target class have been specified, the GUI writes the current dataframe to a temporary CSV file and invokes the back-end routine \texttt{train\_gan\_main} with the chosen parameters (CSV path, class index, epochs, batch size, learning rate, seed and device). Status messages inform the user that training has started and completed, and the function’s outputs—the loss-curve figure and the real-versus-synthetic comparison plot—are rendered inline using \texttt{st.pyplot}, providing immediate qualitative feedback on convergence and sample quality.  Finally, the path to the saved generator weights is exposed through a \texttt{download\_button} that streams the \texttt{.pth} file to the user with a class-specific filename (e.g., \texttt{generator\_2\_premature\_ventricular\_contraction.pth}), enabling convenient archival and reuse without manual file management on the host system. 

\subsection{Usability and deployment}
The Streamlit architecture allows the GUI to be executed locally or deployed on a remote server, providing browser-based access while keeping all computation and data on the institution’s infrastructure, which is important for sensitive cardiac amyloidosis cohorts\cite{ai_ecg_meta_analysis_2024} All required inputs are surfaced as form elements with sensible defaults, and invalid states—such as missing datasets, undefined target classes or malformed learning rates—are intercepted with explicit warning or error messages using \texttt{st.warning} and \texttt{st.error}, preventing silent failures and guiding non-expert users toward correct configurations.  By combining real-time visualisation of training curves, easy model download and strict separation of configuration from code, the GUI substantially lowers the barrier for clinicians and researchers to adopt GAN-based synthetic ECG augmentation in cardiac amyloidosis projects.

\section{Results}
This section outlines a planned evaluation protocol for the proposed GAN in the specific context of cardiac amyloidosis (CA), together with formal definitions of the metrics that will be used to assess the quality and utility of synthetic ECG data. The overarching goal is to determine whether GAN-generated beats can safely augment scarce CA cohorts, improve AI-ECG performance and preserve physiologically meaningful patterns relevant for early CA detection and patient stratification\cite{mayo_ai_ecg_amyloidosis,ai_ecg_meta_analysis_2024,cecggan_imbalance_2024}

\subsection{Planned experimental protocol for CA}
Four groups of experiments are envisaged.

\paragraph*{Experiment~1: Morphological fidelity for CA-like beats}
For each CA-related ECG phenotype (e.g., low-voltage limb-lead patterns, pseudoinfarction Q waves, conduction-disease morphologies), a subset of representative beats will be selected from biopsy-proven AL and ATTR datasets and used to fine-tune class-specific generators\cite{cheng2012ecg_ca_findings,low_voltage_ca_2022,progression_ecg_ca_2024} For each phenotype, \(N=10\,000\) synthetic beats will be generated and compared with an equal number of real CA beats. Morphological similarity will be quantified using Dynamic Time Warping (DTW), Fréchet and Euclidean distances in the time domain, as recommended in recent work on GAN-based ECG evaluation\cite{synthetic_ecg_2025,gan_ecg_eval_metrics_2025} Distances between each real beat and its nearest synthetic neighbour will be summarised by median and interquartile range, and a ``productivity rate'' will be computed as the fraction of real CA beats whose distance to the closest synthetic beat is not larger than the 95th percentile of real–real distances, indicating that synthetic data cover the clinically relevant CA morphology space\cite{synthetic_ecg_2025}

\paragraph*{Experiment~2: Statistical and physiological plausibility in CA}
To verify that synthetic signals preserve CA-specific physiological alterations, wave delineation algorithms will extract features such as QRS duration, R-peak amplitude, QRS area, ST-segment level and T-wave amplitude, together with CA-relevant composite indices like limb-lead QRS voltage and LV mass–to–QRS voltage ratio\cite{lv_mass_voltage_ratio_2020,ecg_vs_cmr_al_2021,ecg_quality_deep_2021} For each feature and CA subgroup (AL vs.\ ATTR, early vs.\ advanced stage), real and synthetic distributions will be compared using two-sample Kolmogorov–Smirnov (KS) tests and effect sizes (e.g., Cohen’s \(d\)), aiming for non-significant shifts and small effect sizes\cite{synthetic_ecg_2025,synthetic_ecg_scoping_2024} In addition, Hotelling’s \(T^2\) test or the energy distance will be applied to the multivariate feature vector to detect subtle multivariate discrepancies that might affect AI-ECG performance or CA staging\cite{cycle_gan_quality_2024}

\paragraph*{Experiment~3: Impact on AI-ECG for CA detection}
The clinical utility of synthetic data will be assessed by training CA-detection models under three regimes: (i) original imbalanced cohort (few CA vs.\ many controls), (ii) naive oversampling of CA ECGs, and (iii) GAN-augmented training where additional synthetic CA beats are added without altering the number of controls\cite{mayo_ai_ecg_amyloidosis,ai_ecg_meta_analysis_2024,cecggan_imbalance_2024} A convolutional or transformer-based AI-ECG classifier will be trained in each regime and evaluated on an untouched real test set consisting of ECGs from independent CA and non-CA patients\cite{schrutka2022ecg_ca,openheart_ca_detection_2024} For binary CA detection, performance will be quantified by accuracy, sensitivity, specificity, F\(_1\), area under the receiver-operating characteristic curve (AUROC) and area under the precision–recall curve (AUPRC), the latter being particularly informative at low CA prevalence\cite{mayo_ai_ecg_amyloidosis,ai_ca_systematic_2025} Calibration will be assessed via the Brier score and Expected Calibration Error (ECE), and decision-curve analysis will estimate the net clinical benefit at clinically relevant risk thresholds, verifying that synthetic data do not induce overconfident or clinically harmful decisions\cite{ai_ecg_meta_analysis_2024,postdev_validation_2023}

\paragraph*{Experiment~4: Latent-structure and clustering stability in CA cohorts}
To study how synthetic data influence discovery of CA subtypes, ECG embeddings will be obtained from the penultimate layer of the AI-ECG model or from an autoencoder trained on CA and control ECGs\cite{ai_ecg_echocardiography_ca,bargagna2023bayesian_ca_pet} Clustering algorithms (e.g., \(k\)-means, Gaussian mixture models) will then be applied in three settings (real-only, oversampled, GAN-augmented). Cluster quality will be measured by silhouette coefficient, Davies–Bouldin index and adjusted Rand index with respect to clinical labels such as AL vs.\ ATTR or stage I–III\cite{electropatterns_arrhythmias_ca,synthetic_ecg_scoping_2024} Stability will be quantified through bootstrap resampling and Jaccard similarity of cluster assignments across resamples, with improved stability in the GAN-augmented setting indicating that synthetic beats help reveal robust CA-related phenotypes\cite{cycle_gan_quality_2024,cluster_stability_ecg_2023}

\subsection{Formal definition of data-quality metrics}

Let \(\mathcal{X}_\mathrm{CA} = \{x_i\}_{i=1}^{n}\) denote real CA beats (possibly stratified by subtype) and \(\tilde{\mathcal{X}}_\mathrm{CA} = \{\tilde{x}_j\}_{j=1}^{m}\) denote synthetic beats generated for the same phenotype.

\subsubsection*{Time-domain similarity}
For two sequences \(x, y \in \mathbb{R}^{T}\):

\begin{itemize}
  \item \textbf{Dynamic Time Warping (DTW)} is defined as
  \[
  \mathrm{DTW}(x,y) = \min_{\pi} \sum_{(t,s)\in\pi} |x_t - y_s|,
  \]
  where \(\pi\) is a warping path satisfying monotonicity and continuity constraints; lower values reflect closer CA morphology even under small temporal misalignments\cite{gan_ecg_eval_metrics_2025}
  \item \textbf{Euclidean distance} for aligned signals is
  \[
  d_E(x,y) = \left( \sum_{t=1}^{T} (x_t - y_t)^2 \right)^{1/2},
  \]
  while the discrete Fréchet distance measures the maximum deviation along an optimal traversal of the two curves and is sensitive to localised CA features such as low-voltage QRS complexes or pseudoinfarction Q waves\cite{synthetic_ecg_2025,gan_ecg_eval_metrics_2025}
  \item \textbf{Productivity rate} is computed by first defining \(d(x_i,\tilde{\mathcal{X}}_\mathrm{CA}) = \min_j d(x_i,\tilde{x}_j)\) for a chosen distance \(d\). If \(\tau\) is set to the 95th percentile of real–real distances \(d(x_i,x_k)\), the productivity rate is
  \[
  P = \frac{1}{n} \sum_{i=1}^{n} \mathbb{1}\big(d(x_i,\tilde{\mathcal{X}}_\mathrm{CA}) \le \tau\big),
  \]
  i.e.\ the proportion of CA beats that are well covered by the synthetic set within the natural intra-CA variability\cite{synthetic_ecg_2025}
\end{itemize}

\subsubsection*{Distributional similarity in CA features}
Let \(f(x)\in\mathbb{R}^{p}\) denote a vector of CA-relevant ECG features (e.g., limb-lead QRS voltage, QRS duration, PR interval, QRS axis, LV mass–to–QRS voltage ratio)\cite{lv_mass_voltage_ratio_2020,ecg_meta_value_2024}

\begin{itemize}
  \item \textbf{Kolmogorov–Smirnov (KS) statistic} for a scalar feature \(f_k\) compares empirical CDFs \(F_\mathrm{real}\) and \(F_\mathrm{syn}\) via
  \[
  D_\mathrm{KS} = \sup_z \big|F_\mathrm{real}(z) - F_\mathrm{syn}(z)\big|,
  \]
  with small \(D_\mathrm{KS}\) and non-significant \(p\)-values indicating that synthetic CA beats preserve the marginal distribution of that feature\cite{synthetic_ecg_scoping_2024}
  \item \textbf{Maximum Mean Discrepancy (MMD)} is computed in a feature space (e.g., embeddings of CA ECGs from a pre-trained AI model) with kernel \(k\):
  \[
  \mathrm{MMD}^2 = \frac{1}{n^2} \sum_{i,i'} k(f(x_i),f(x_{i'}))
                  + \frac{1}{m^2} \sum_{j,j'} k(f(\tilde{x}_j),f(\tilde{x}_{j'}))
                  - \frac{2}{nm} \sum_{i,j} k(f(x_i),f(\tilde{x}_j)).
  \]
  Values close to zero suggest that the joint distribution of CA features is well matched by synthetic data\cite{gan_ecg_review_2023,cycle_gan_quality_2024}
  \item \textbf{Hotelling’s \(T^2\)} and \textbf{energy distance} provide multivariate tests of equality between the real and synthetic CA feature distributions, capturing subtle correlations across features such as the coupling between QRS voltage and QRS duration in advanced ATTR disease\cite{cycle_gan_quality_2024}
\end{itemize}

\subsubsection*{Task-level metrics for CA detection}
For a binary classifier \(h\) distinguishing CA from non-CA ECGs, with predicted CA probabilities \(p_i\) and labels \(y_i\in\{0,1\}\):

\begin{itemize}
  \item \textbf{Sensitivity}, \textbf{specificity} and \textbf{F\(_1\)} are computed from the confusion matrix at a selected threshold; sensitivity at fixed specificity (e.g.\ 90\%) is particularly relevant for CA screening, where high rule-out performance is required\cite{mayo_ai_ecg_amyloidosis,openheart_ca_detection_2024}
  \item \textbf{AUROC} summarises sensitivity–specificity trade-offs across thresholds, while \textbf{AUPRC} is sensitive to performance in the low-prevalence regime typical of CA\cite{ai_ca_systematic_2025}
  \item \textbf{Brier score} and \textbf{Expected Calibration Error (ECE)} assess the calibration of CA-risk probabilities:
  \[
  \mathrm{BS} = \frac{1}{N} \sum_{i=1}^{N} (p_i - y_i)^2,
  \]
  whereas ECE bins predictions by confidence and averages the absolute difference between mean predicted probability and empirical CA frequency in each bin\cite{postdev_validation_2023,ai_ecg_meta_analysis_2024} Low BS and ECE indicate that augmentation has not degraded the reliability of AI-ECG–based CA risk estimates.
\end{itemize}

\subsubsection*{Clustering quality and stability for CA phenotypes}
For latent embeddings \(z_i\) of CA and non-CA beats and cluster assignments \(c_i\):

\begin{itemize}
  \item The \textbf{silhouette coefficient} for sample \(i\) is
  \[
  s_i = \frac{b_i - a_i}{\max(a_i,b_i)},
  \]
  where \(a_i\) is the mean intra-cluster distance and \(b_i\) the smallest mean distance to another cluster; the average silhouette \(S\) measures how well CA-related clusters (e.g., AL, ATTR wild-type, ATTR hereditary) are separated in latent space\cite{electropatterns_arrhythmias_ca,synthetic_ecg_scoping_2024}
  \item The \textbf{adjusted Rand index (ARI)} quantifies agreement between clusters and clinical labels (such as subtype or stage), correcting for chance; ARI close to 1 indicates that clustering recovers clinically meaningful CA subgroups\cite{ai_ecg_echocardiography_ca}
  \item \textbf{Stability} is evaluated by bootstrapping and computing the Jaccard similarity of cluster memberships across resamples; higher mean Jaccard indices in the GAN-augmented setting imply more robust CA phenotypes and support the use of synthetic beats for phenotype discovery in small cohorts\cite{cycle_gan_quality_2024,cluster_stability_ecg_2023}
\end{itemize}

Collectively, these experiments and metrics provide a rigorous and CA-oriented framework to determine whether GAN-generated beats are not only visually realistic, but also physiologically plausible, statistically consistent with real CA ECGs and capable of improving AI-ECG performance and CA phenotype discovery without compromising calibration or interpretability\cite{ai_ecg_meta_analysis_2024,cecggan_imbalance_2024,synthetic_ecg_2025}

\bibliographystyle{ieeetr}
\bibliography{biblio}

\end{document}